\documentclass[10pt,letterpaper]{article}
\usepackage[top=0.85in,left=2.75in,footskip=0.75in]{geometry}

\usepackage{amsmath,amssymb}

\usepackage{changepage}

\usepackage{textcomp,marvosym}

\usepackage{cite}

\usepackage{nameref,hyperref}

\usepackage[right]{lineno}

\usepackage[nopatch=eqnum]{microtype}
\DisableLigatures[f]{encoding = *, family = * }

\usepackage[table]{xcolor}

\usepackage{array}

\usepackage{subcaption}

\newcolumntype{+}{!{\vrule width 2pt}}

\newlength\savedwidth

\raggedright
\usepackage[aboveskip=1pt,labelfont=bf,labelsep=period,justification=raggedright,singlelinecheck=off]{caption}

\makeatletter
\renewcommand{\@biblabel}[1]{\quad#1.}
\makeatother

\usepackage{lastpage,fancyhdr,graphicx}
\usepackage{epstopdf}
\fancyheadoffset[L]{2.25in}
\fancyfootoffset[L]{2.25in}
\begin{document}
\vspace*{0.2in}

\begin{flushleft}
{\Large
\textbf\newline{Smartphone region-wise image indoor localization using deep learning for indoor tourist attraction} 
}
\newline
\\
Gabriel Toshio Hirokawa Higa\textsuperscript{1*\textcurrency},
Rodrigo Stuqui Monzani\textsuperscript{2},
Jorge Fernando da Silva Cecatto\textsuperscript{2},
Maria Fernanda Balestieri Mariano de Souza\textsuperscript{3},
Vanessa Aparecida de Moraes Weber\textsuperscript{4, 5},
Hemerson Pistori\textsuperscript{1, 2},
Edson Takashi Matsubara\textsuperscript{2},
\\
\bigskip
\textbf{1} Dom Bosco Catholic University, Campo Grande, MS, Brazil
\\
\textbf{2} Federal University of Mato Grosso do Sul, Campo Grande, MS, Brazil
\\
\textbf{3} Pantanal Biopark, Campo Grande, MS, Brazil
\\
\textbf{4} State University of Mato Grosso do Sul, Campo Grande, MS, Brazil
\\
\textbf{5} Kerow Precision Solutions, Campo Grande, MS, Brazil
\\
\bigskip

%
%


\textcurrency Current Address: Dom Bosco Catholic University, Campo Grande, MS, Brazil 



* gabrieltoshio03@gmail.com

\end{flushleft}
\section*{Abstract}
Smart indoor tourist attractions, such as smart museums and aquariums, usually require a significant investment in indoor localization devices. The smartphone Global Positional Systems use is unsuitable for scenarios where dense materials such as concrete and metal block weaken the GPS signals, which is the most common scenario in an indoor tourist attraction. Deep learning makes it possible to perform region-wise indoor localization using smartphone images. This approach does not require any investment in infrastructure, reducing the cost and time to turn museums and aquariums into smart museums or smart aquariums. This paper proposes using deep learning algorithms to classify locations using smartphone camera images for indoor tourism attractions. We evaluate our proposal in a real-world scenario in Brazil. We extensively collect images from ten different smartphones to classify biome-themed fish tanks inside the Pantanal Biopark, creating a new dataset of 3654 images. We tested seven state-of-the-art neural networks, three being transformer-based, achieving precision around 90\% on average and recall and f-score around 89\% on average. The results indicate good feasibility of the proposal in a most indoor tourist attractions.



\section*{Introduction}
The Pantanal Biopark\footnote{Biopark Pantanal, as the official name goes in Portuguese.} is a state public aquarium located in Campo Grande (20° 45' 60.189" S latitude and $-$54° 58' 07.123" W longitude), Mato Grosso do Sul, Brazil. It is the largest freshwater aquarium in the world, with 19 thousand m$^2$ of built area and more than 5 million liters of water in 93 tanks, 70 of which are dedicated to research and conservation. Over 359 species of animals from the Pantanal and other regions are present in thematic tanks, offering visitors the experience of the main Brazilian ecosystems, including chimney springs, floodplains, and footpaths.

In addition to the purpose of tourism, the Biopark was designed as an educational environment, within which new tools shall be developed and made available for visitors to use, incorporating state-of-the-art technologies, especially those from the Artificial Intelligence (AI) field, into the learning experience. One of these tools is a smartphone app that automatically identifies the thematic tank the visitor is pointing to the camera. The image classification, where the classes represent fish tank locations,  enables an essential function of this application to allow the visitors to locate themselves inside the Biopark. 

Indoor localization inside Biopark using mobile phones is a challenging task. This problem is familiar to most indoor tourist attractions since, in most cases, these places are built with metallic structures and concrete due to their architectural construction. Traditional GPS methods are inaccurate inside such places since the structures block the GPS signal. There are several approaches to solving this type of problem, the most famous among them being the location by wireless technologies, such as Wi-Fi and Bluetooth. However, physical resources would still be needed to allow the location of visitors. Therefore, a cheaper alternative is to develop technologies that do not require the installation of any infrastructure but use the technology already available. Our proposal explore a well-known approach to region-wise localization by using image classification. The main idea lies in the combination of image classification of the fish tank and prior knowledge of the arrangement of tanks in the park, which enables a simple yet effective indoor localization system.

The main contributions of this paper are:

\begin{itemize}
    \item A novel image dataset with 3654 images from 24 different fish tanks (locations), 23 indoor locations, and one outdoor location of  Biopark Pantanal;
    \item  An experimental evaluation of seven deep learning algorithms, where we focused on the proposal's applicability by selecting small and medium-size models that anyone with a standard GPU can train. Also, the small models can fit on any standard mobile phone device; and
    \item A discussion of how an indoor localization system based on the new dataset shall work.
\end{itemize}

\section*{Related works}
\label{Related works}
\begin{figure}[!htb]
    \centering
    \includegraphics[width=0.4\textwidth]{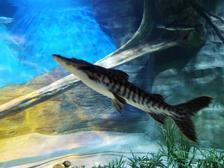}
    \caption{a fish in the \textit{Banhados} tank.}
    \label{fig:exemplo}
\end{figure}

The goal of the learned model in this paper is to consider as an input an image and the output is the location (fish tank name), which at Biopark Pantanal refers to the biomes that the tanks purport to mimic. The complex backgrounds and the fish species are central features of the classification. It is also worth noticing that images captured underwater through a glass panel, as is the case with images captured in an aquarium, present their peculiarity, such as light distortions because of light coming through the surface of the water (see, for instance, Figure~\ref{fig:exemplo}). Therefore related works to this study include fish tank classification, fish classification, and indoor localization using deep learning.

Ubina \textit{et al.}~\cite{ubina_a_visual_2021} addressed fish tank detection with deep learning techniques. In their work, the authors created a complex system using drones to automate different tasks (not only fish-related ones but also security surveillance tasks) in fish aquaculture sites. Their system did include the detection of aquaculture tanks using deep learning. Their problem was, however, clearly very different from the one addressed in this work, not only because of how the images were captured but also because the tanks were only classified amongst other kinds of objects (such as people and boats), and not amongst themselves, as tanks diversely themed according to different aquatic environments. 

Most research on the application of deep learning in fish-related sciences up until now has focused on underwater images captured in marine environments. Zhao \textit{et al.}~\cite{zhao_application_2021} reviewed the application of machine learning techniques in the field of intelligent aquaculture, indicating that there are good results in the task of fish species classification (with an accuracy usually above 90\%, frequently reaching 98\%, and even going above 99\% in at least one case reported in their work), but also that these results are highly dependent on the dataset and on the environmental conditions. The preferred datasets are the Fish4Knowledge dataset and versions of the LifeCLEF dataset. Many works address the complexity of the environments. Salman \textit{et al.}~\cite{salman_automatic_2020}, for example, addressed the issue of complex backgrounds (originating from complex environments) in underwater video processing by using a hybrid model consisting of gaussian mixture modeling and optical flow. Ju and Xue~\cite{ju_fish_2020} also addressed the problem of complex backgrounds by proposing the Fish\_AlexNet, a neural network based on the AlexNet, improved with an item-based soft attention mechanism. These works show that fish image classification is now a well-established field addressing its difficulties in their specificities. 

Looking at indoor localization literature,  Félix \textit{et al.}~\cite{felix2016fingerprinting} proposed in 2016 the use of deep learning on indoor localization. The authors evaluate the fingerprinting of the Wi-FI access point. Since 2016, deep learning models have evolved immensely, and now deep learning for mobile devices is a very feasible approach. Shao \textit{et al.}~\cite{shao2018indoor} transformed Wi-Fi and magnetic field fingerprints into an image-like structure and processed it with a convolutional neural network. Their result shows an impressive under one-meter error distance using different smartphone orientations. Although the result is satisfying, they used Wi-Fi and magnetic fields as the primary input, which is not our case.

Liu \textit{et al.}~\cite{liu2020indoor} proposed a system for indoor localization that uses both place and object recognition in images to map pictures into environments. Their primary input uses RGB images from the camera, where  SURF, FLANN, and RANSAC preprocess the images to identify keyframes. The proposal submits the keyframes to a Deep Learning Object detection and performs the localization. Although the results are promising, their proposal requires server-client architecture to process all images due to its computational demands. 

Bai \textit{et al.}~\cite{bai2019survey} presents a survey on deep learning image-based approach. Most works at that time do not implement deep learning end-to-end approaches but the combination of traditional computer vision methods (ex. SIFT, SURF, BOVW)  with deep learning. Another interesting observation in this review is that most papers perform a multi-fusion of an image, WIFI fingerprinting, gyroscope, and accelerometer data. 

The problem of classifying tanks seems to be an open question. Few steps have been taken lately toward its solution, and there are good reasons to speed up. Being in a different context, Sergi \textit{et al.}~\cite{sergi_a_microservices_2022} proposed a system that is related to ours in its spirit: enhancing educational tourism and learning experiences with deep learning. In their case, the proposed system used image-matching deep learning techniques to enhance the fruition of cultural tourism. The system can display information on cultural points of interest by matching user input into reference images stored in the system. In the case of this work, the techniques use image classification. As the authors say, it is possible to enhance the experience of learning about culture with artificial intelligence tools. The same can be said about biomes and aquatic fauna: it is possible to enhance the experience of learning about them with AI. Of course, the correct identification of biome-themed fish tanks is necessary. Therefore, in this step, we contribute by presenting an artificial neural network to classify biome-themed fish tank images.

\section*{Proposal}
\label{proposal}

The proposal follows a standard pipeline of a fingerprinting localization problem: it starts with an image capture, followed by a deep learning image classification to identify the fish tank, and finally, the match of the fish tank name and the localization. By accessing an indoor localization system based on smartphone cameras, a user must be able to input a picture of the tank and get their localization as an answer. For example, for Figure~\ref{fig:exemplo}, the system should inform the user that they are looking at the Wetlands-themed tank. The system may also return additional information and trivia about the tank and the species. Although this approach has already been used in many different applications, to the best of our knowledge, our study is the first to evaluate this approach in a fish tank aquarium, where specific indoor light and environmental conditions are presented, in which we believe that can generalizes in other similar indoor tourist attraction.

In this work, we consider two ways to process the input data: online and offline. For online processing, options vary from having an actual server working in the Biopark to using cloud computing; therefore, we can use GPUs and bigger networks. For offline processing, it is possible to embed a smaller model through the smartphone application and run it in the user's smartphone. These options require that many factors be considered, especially in this paper's context, the performance of smaller networks that can run in smartphones in comparison with larger neural networks, which require more computational power and may achieve better performance.

In principle, using the aforementioned models, it is possible to identify a tank from an image captured by a smartphone. In addition, it is also possible to observe that even networks with a smaller number of parameters present acceptable results for this task. These facts support creating a mobile application that can work online, making requests to an external network allocated on a server or offline since it is possible to embed the networks presented in an application.

\section*{Materials and methods}
\label{methods}
\subsection*{Dataset}

\begin{figure*}[!htb]
    \centering
    \includegraphics[width=1\textwidth]{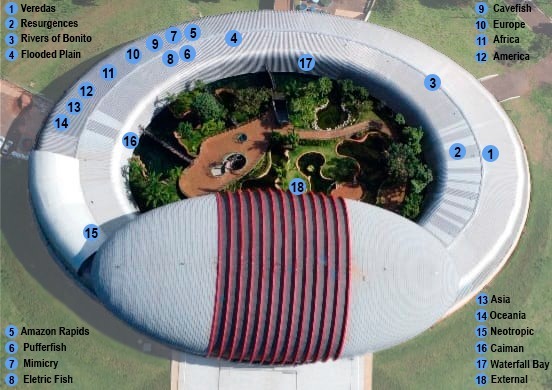}
    \caption{A 2D map of Pantanal Biopark}
    \label{fig:map}
\end{figure*}

The dataset classes are based on the Pantanal Biopark tank division. The tanks represented in the dataset are: Africa, America, Asia, Waterfall Bay, Pufferfish, Wetlands, Amazon Rapids, Europe, Creeks, Caiman, Australian Lake, Mimicry, Neotropic, Oceania, Electric Fish, Flooded Plain, Flooded Plain (Flooding), Resurgence, Rivers of Bonito, Rivers of the Pantanal, Veredas and External.  Figure~\ref{fig:map} shows the map of the fish tank used in this proposal. 

\begin{figure*}
  \centering
  \begin{subfigure}{0.2\textwidth}
    \includegraphics[width=1\linewidth,height=0.75\linewidth]{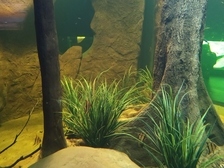} 
    \caption{Africa}
    \label{fig:mosaico_africa}
  \end{subfigure}
  \quad
  \begin{subfigure}{0.2\textwidth}
    \includegraphics[width=1\linewidth,height=0.75\linewidth]{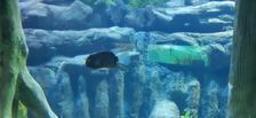} 
    \caption{America}
    \label{fig:mosaico_america}
  \end{subfigure}
  \quad
  \begin{subfigure}{0.2\textwidth}
    \includegraphics[width=1\linewidth,height=0.75\linewidth]{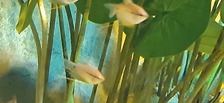} 
    \caption{Asia}
    \label{fig:mosaico_asia}
  \end{subfigure}
  \quad
  \begin{subfigure}{0.2\textwidth}
    \includegraphics[width=1\linewidth,height=0.75\linewidth]{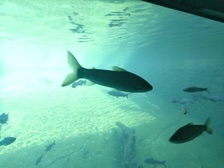} 
    \caption{Waterfall bay}
    \label{fig:mosaico_baia_cachoeira}
  \end{subfigure}
  \quad
  \begin{subfigure}{0.2\textwidth}
    \includegraphics[width=1\linewidth,height=0.75\linewidth]{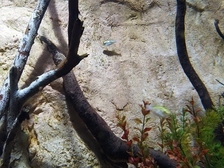} 
    \caption{Pufferfish}
    \label{fig:mosaico_baiacus}
  \end{subfigure}
  \quad
  \begin{subfigure}{0.2\textwidth}
    \includegraphics[width=1\linewidth,height=0.75\linewidth]{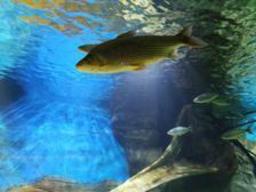} 
    \caption{Wetlands}
    \label{fig:mosaico_banhados}
  \end{subfigure}
  \quad
  \begin{subfigure}{0.2\textwidth}
    \includegraphics[width=1\linewidth,height=0.75\linewidth]{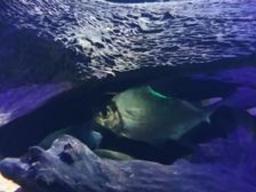} 
    \caption{Amazon rapids}
    \label{fig:mosaico_corr_amazonia}
  \end{subfigure}
  \quad
  \begin{subfigure}{0.2\textwidth}
    \includegraphics[width=1\linewidth,height=0.75\linewidth]{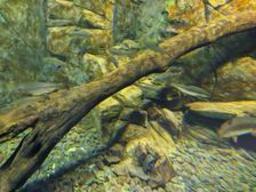} 
    \caption{Europe}
    \label{fig:mosaico_europa}
  \end{subfigure}
  \quad
  \begin{subfigure}{0.2\textwidth}
    \includegraphics[width=1\linewidth,height=0.75\linewidth]{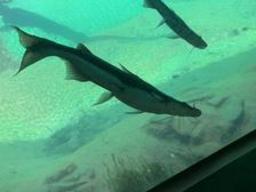} 
    \caption{External}
    \label{fig:mosaico_externo}
  \end{subfigure}
  \quad
  \begin{subfigure}{0.2\textwidth}
    \includegraphics[width=1\linewidth,height=0.75\linewidth]{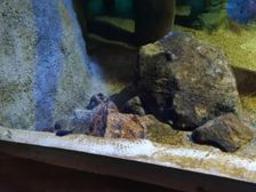} 
    \caption{Creeks}
    \label{fig:mosaico_igarapes}
  \end{subfigure}
  \quad
  \begin{subfigure}{0.2\textwidth}
    \includegraphics[width=1\linewidth,height=0.75\linewidth]{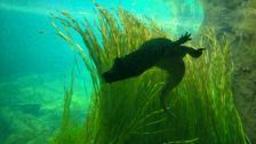} 
    \caption{Caiman}
    \label{fig:mosaico_jacares}
  \end{subfigure}
  \quad
  \begin{subfigure}{0.2\textwidth}
    \includegraphics[width=1\linewidth,height=0.75\linewidth]{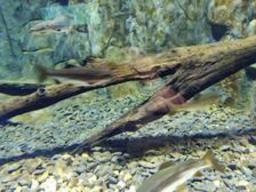} 
    \caption{Australian Lake}
    \label{fig:mosaico_lagoa_australiana}
  \end{subfigure}
  \quad
  \begin{subfigure}{0.2\textwidth}
    \includegraphics[width=1\linewidth,height=0.75\linewidth]{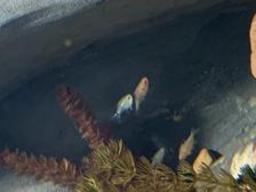} 
    \caption{Mimicry}
    \label{fig:mosaico_mimetismo}
  \end{subfigure}
  \quad
  \begin{subfigure}{0.2\textwidth}
    \includegraphics[width=1\linewidth,height=0.75\linewidth]{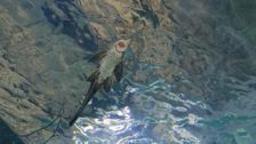} 
    \caption{Neotropic}
    \label{fig:mosaico_neotropico}
  \end{subfigure}
  \quad
  \begin{subfigure}{0.2\textwidth}
    \includegraphics[width=1\linewidth,height=0.75\linewidth]{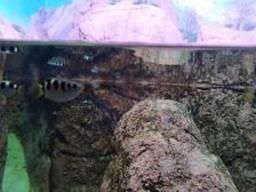} 
    \caption{Oceania}
    \label{fig:mosaico_oceania}
  \end{subfigure}
  \quad
  \begin{subfigure}{0.2\textwidth}
    \includegraphics[width=1\linewidth,height=0.75\linewidth]{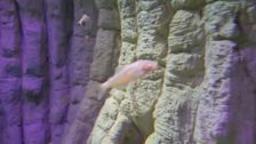} 
    \caption{Cavefish}
    \label{fig:mosaico_cavernicolas}
  \end{subfigure}
  \quad
  \begin{subfigure}{0.2\textwidth}
    \includegraphics[width=1\linewidth,height=0.75\linewidth]{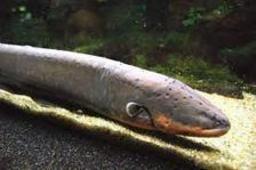} 
    \caption{Electric fish}
    \label{fig:mosaico_peixe_eletrico}
  \end{subfigure}
  \quad
  \begin{subfigure}{0.2\textwidth}
    \includegraphics[width=1\linewidth,height=0.75\linewidth]{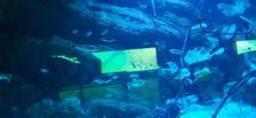} 
    \caption{Flooded plain}
    \label{fig:mosaico_plan_inundada}
  \end{subfigure}
  \quad
  \begin{subfigure}{0.2\textwidth}
    \includegraphics[width=1\linewidth,height=0.75\linewidth]{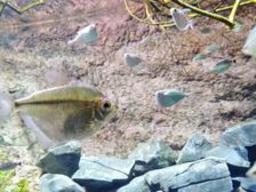} 
    \caption{Flooded plain (flooding)}
    \label{fig:mosaico_plan_inundadacheia}
  \end{subfigure}
  \quad
  \begin{subfigure}{0.2\textwidth}
    \includegraphics[width=1\linewidth,height=0.75\linewidth]{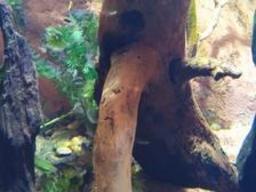} 
    \caption{Flooded plain (dry)}
    \label{fig:mosaico_plan_inundadaseca}
  \end{subfigure}
  \quad
  \begin{subfigure}{0.2\textwidth}
    \includegraphics[width=1\linewidth,height=0.75\linewidth]{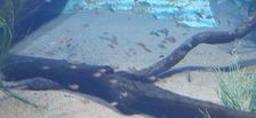} 
    \caption{Resurgences}
    \label{fig:mosaico_ressurgencias}
  \end{subfigure}
  \quad
  \begin{subfigure}{0.2\textwidth}
    \includegraphics[width=1\linewidth,height=0.75\linewidth]{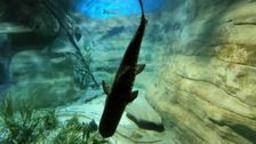} 
    \caption{Rivers of Bonito}
    \label{fig:mosaico_rios_bonito}
  \end{subfigure}
  \quad
  \begin{subfigure}{0.2\textwidth}
    \includegraphics[width=1\linewidth,height=0.75\linewidth]{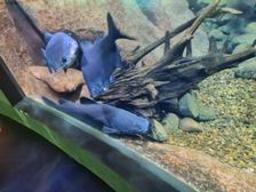} 
    \caption{Rivers of the Pantanal}
    \label{fig:mosaico_rios_pantanal}
  \end{subfigure}
  \quad
  \begin{subfigure}{0.2\textwidth}
    \includegraphics[width=1\linewidth,height=0.75\linewidth]{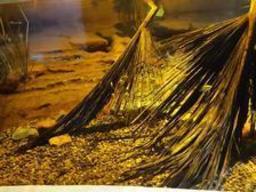} 
    \caption{Veredas}
    \label{fig:mosaico_veredas}
  \end{subfigure}
  \quad

   \caption{One image sample from each tank used in the experiment.}
   \label{fig:tanks}
\end{figure*}

The images from the tanks were taken on May 21st, 2022, at 2 pm, by a group of 9 students from the Federal University of Mato Grosso do Sul (UFMS) in a tour guided by Prof. Edson Takashi Matsubara, who also contributed to the collection. The students were instructed to use their smartphones to collect as many images as possible with different angles and light conditions. In addition, it was also requested that there be a balance between the number of images recorded in each tank to avoid imbalance between classes. Table~\ref{table:smartphones} shows the smartphone models, the resolutions of their cameras, and the number of images taken using each one, both in total and percentage. From the 31 tanks available, only 24 were used in this work, as the tanks with fewer than 50 images were removed because the deep learning techniques required more images to learn. In total, there are 3654 images of 24 tanks. Table~\ref{table:tanks} shows the number of images of each tank. In Figure~\ref{fig:tanks}, it is possible to see one sample image from each tank.

Among the main species of each habitat, the following stand out:
\begin{itemize}
    \item Africa: african cichlid, petricola catfish (\textit{Synodontis petricola}) and featherfin chichlid (\textit{Cyathopharynx furcifer});
    
    \item America: triangle cichlid waroo (\textit{Uaru amphiacanthoides}), porcupinefish (\textit{Geophagus spp.}), true parrotfish (\textit{Hoplarchus psittacus}),bandit cichlid (\textit{Guianacara dacrya}), armored catfish, serverum (\textit{Heros severus}), silver dollar (\textit{Leporinus}) and pike cichlid (\textit{Crenicichla spp.});
    
    \item Asia:  barb, clown loach (\textit{Chromobotia macracanthus}), Hara jerdoni (Botia kubotai), Black shark minnow (Epalzeorhynchos bicolor), Three-striped corydoras (Trichopodus leerii), Aurelius Barb (Dawkinsia arulius), Golden barb (Barbodes semifasciolatus) and Yo-yo loach (Botia lohachata);
        
    \item Waterfall Bay: piraputanga (\textit{Brycon hilarii}), armored catfish, Pantanal catfish (\textit{Pimelodus pantaneiro}), swamp eel (\textit{Synbranchus spp.}), schizodon (\textit{Schizodon borellii}), silver dollar (\textit{Leporinus friderici}), silver prochilod (\textit{Mylossoma duriventre}), large-headed leporinus (\textit{Megaleporinus macrocephalus}), and tortoise;
    
    \item Pufferfish: Amazon pufferfish (\textit{Colomesus asellus}), greenbottle pufferfish (\textit{Auriglobus nefastus});
    
    \item Wetlands: piraputanga (\textit{Brycon hilarii}), curimata-serrasalmid (\textit{Prochilodus lineatus}), thicklip cichlid (\textit{Cichasoma dimerus}), spotted pike cichlid (\textit{Crenicichla lepidota}) and African butterfly fish (\textit{Markiana nigripinnis});
    
    \item Amazon Rapids: zebra pleco (\textit{Hypancistrus zebra}), Hypancistrus catfish (\textit{Hypancistrus sp.}), golden nugget pleco (\textit{Baryancistrus xanthellus}), Baryancistrus catfish, cactus plecos (\textit{Pseudacanthicus spp.}), sailfin plecos (\textit{Leporacanthicus spp.}), flagtail prochilod (\textit{Acnodon Normani}) and bloodfin tetra (\textit{Moenkhausia heikoi});
    
    \item Europe: rainbow trout (\textit{Oncorhynchus mykiss});
    
    \item Creeks: Panda corydoras (\textit{Corydoras panda}), Neon tetra (\textit{Paracheirodon innesi}), Green neon tetra (\textit{Paracheirodon simulans}), Glowlight tetra (\textit{Hemigrammus erythrozonus}) Nego d’água (\textit{Hyphessobrycon negodagua}), Diamond tetra (\textit{Moenkhausia pittieri}), Tetra Peugeot (\textit{Hyphessobrycon Peugeot}), Lemon tetra (\textit{Hyphessobrycon pulchripinnis}) Penguin tetra (\textit{Thayeria boehlkei}), Costae tetra (\textit{Moenkhausia costae}), Rummy-nose tetra (\textit{Hemigrammus bleheri}), Cardinal tetra (\textit{Paracheirodon axelrodi}), Spotted headstander (\textit{Chilodus punctatus}), Cruzeiro do Sul (\textit{Hemiodus gracilis}), Altum angelfish (\textit{Pterophyllum altum}) and Discus fish (\textit{Symphysodon discus});
    
    \item Caiman:  broad-snouted caiman (\textit{Caiman Latirostris}), black belt cichlid (\textit{Aequidens plagiozonatus}), flag acara (\textit{Bujurquina vittata}), thicklip cichlid (\textit{Cichlasoma dimerus}), black tetra (\textit{Moenkhausia dichroura}) and silver dollar fish (\textit{Tetragonopterus argenteus});
    
    \item Australian Lake: Rainbowfish;
    
    \item Mimicry: Dwarf corydoras (\textit{Corydoras hastatus}), Lips (\textit{Hemigrammus mahnerti}), Krieg's tetra (\textit{Serrapinnus kriegi}), Serpae tetra (\textit{Hyphessobrycon eques}), Black Phantom tetra (\textit{Hyphessobrycon megalopterus}), Lips (\textit{Hyphessobrycn elachys}) and Armored catfish;
    
    \item Neotropic: Brazilian dourado (\textit{Salminus brasiliensis}), stingray, pintado (\textit{Pseudoplatystoma corruscans}), Tambaqui (\textit{Colossoma macropomum}), thorny catfish (\textit{Oxydoras kneri}), Astyanax (\textit{Astyanax lacustres}), spine-bellied jupiaba (\textit{Jupiaba acanthogaster}), Pirarucu (\textit{Arapaima gigas}), disk tetra (\textit{Myleus schomburkii}), armored catfish, silver dollar, red head tapajos (\textit{Geophagus spp.}), silver arowana (\textit{Osteoglossum bicirrhosum}), redtail catfish (\textit{Phractocephalus hemeliopterus}) and pike cichlid (\textit{Crenicichla spp.});
    
    \item Oceania: archerfish (\textit{Toxotes jacutrix}), archerfish (\textit{Toxotes jaculatrix}), fingerfish (\textit{Monodactylus sabae}) and silver moony (\textit{Monodactylus argenteus});
    
    \item Electric Fish: Black ghost knifefish (\textit{Apteronotus spp.});
    
    \item Flooded Plain: Marbled headstander (\textit{Abramites hypselonotus}), Armored catfish, stingray, Canivete (\textit{Leporinus striatus}), Catfish (\textit{Auchenipterus osteomystax}), serrasalmid (\textit{Myloplus levis}), Pejerrey (\textit{Schizodon isognathus}), Black tetra (\textit{Gymnocorymbus ternetzi}), Argentine humphead (\textit{Gymnogeophagus balzanii}), Astyanax (\textit{Astyanax alleni}), Sardinha (\textit{Triportheus pantanensis}) and whiptail armored catfish (\textit{Loricaria spp.});
    
    \item Flooded Plain (Flooding): papudinho (\textit{Poptella paraguayensis}), Astyanax (\textit{Psalidodon marionae}), flag cichlid (\textit{Mesonauta festivus}) and Redeye tetra (\textit{Moenkhausia forestii});
    
    \item Resurgences: jewel tetra (\textit{Hyphessobrycon eques}), whiptail armored catfish, tetra (\textit{Moenkhausia bonita}) and saguiru (\textit{Toothless characin}); 
    
    \item Rivers of Bonito: Piraputanga (\textit{Brycon hilarii}), thicklip cichlid (\textit{Cichlasoma dimerus}), Pike cichlid (\textit{Crenicichla lepidota}), Armored catfish, stingray, ray-finned fish (\textit{Prochilodus lineatus}), Astyanax (\textit{Psalidodon marionae}, \textit{Jupiaba acanthogaster}, \textit{Hyphessobrycon luetkeni}, \textit{Moenkhausia bonita}), Canivete (\textit{Leporellus vittatus}, \textit{Leporinus striatus}), Duro-duro (\textit{Parodon nasus}), Headstander (\textit{Megaleporinus obtusidens}), stingray and tiger shovelnose catfish (\textit{Pseudoplatysto reticulatum});
    
    \item Rivers of the Pantanal: thorny catfish (\textit{Oxydoras kneri}), stingray (\textit{Potamotrygon spp.}), Armored catfish, Gilded catfish (\textit{Zungaru jahu}), Brazilian dourado (\textit{Salminus brasiliensis}), serrasalmid (\textit{Piaractus mesopotamicus}), pike characin (\textit{Acestrorhynchus pantaneiro}) and Pterodoras granulosus (\textit{Pterodoras granulosus});
    
    \item Veredas: Armored catfish, Astyanax (\textit{Astyanax sp.}), Tetra (\textit{Hyphessobrycon langeani}), Angelfish (\textit{Aequidens sp.}), Chum-chum (\textit{Pimelodella spp.}) and Silver Dollar (\textit{Leporinus sp.});
    
    \item External: Armored catfish, Jurupensém (\textit{Sorubim lima}), Oscar (\textit{Astronotus crassipinnis}), serrasalmid (\textit{Metynnis mola}), Sardinha (\textit{Triportheus pantanensis}), headstanders (\textit{Leporinus lacustris}) and Pike cichlid (\textit{Crenicichla vittata});

\end{itemize}

\begin{table*}[h!]
\centering
\caption{Description of each smartphone used to build the dataset, including the number and percentage of images.}
\label{table:smartphones}
\begin{tabular}{|c|c|c|c|} 
   \hline
   Model    & Specification          & \# Images                  & Percentage \\ 
   \hline\hline
   Apple iPhone XR & 12 Mp & 264 & 7.22\%\\
   \hline
   Samsung S20+ & 12 Mp/64 Mp/12 Mp & 316 & 8.65\%\\
   \hline
   Realme 7 Pro & 64 Mp/8 Mp/2 Mp/2 Mp & 181 & 4.95\%\\
   \hline
   Xiaomi Pocophone F1 & 12 Mp/5 Mp & 109 & 2.98\%\\
   \hline
   Redmi Note 8 & 48 Mp/8 Mp/2 Mp/2 Mp & 148 & 4.05\%\\
   \hline
   Samsung A30 & 16 Mp/5 Mp & 316 & 8.65\%\\  
   \hline
   Samsung S7 & 12 Mp & 627 & 17.16\%\\
   \hline
   Samsung S22 & 50 Mp/12 Mp/10 Mp & 481 & 13.16\%\\
   \hline
   Xiaomi Mi 8 Lite & 12 Mp/5 Mp & 829 & 22.69\%\\
   \hline
   Samsung M51 & 64 Mp/12 Mp/5 Mp/5 Mp & 383 & 10.48\%\\
   \hline
\end{tabular}%
\end{table*}

\begin{table}[!htbp]
\centering
\caption{Number and percentage of images from each tank used in the experiment. The table also shows a brief description of the biome that the tank represents.}
\label{table:tanks}
\resizebox{0.6\columnwidth}{!}{%
\begin{tabular}[h]{|c|c|c|}
    \hline
    Tanks &  \# Images & Percentage\\
    \hline\hline
    Africa & 265 & 7.25\%\\
    \hline
    America & 196 & 5.36\%\\
    \hline
    Asia & 61 & 1.67\%\\
    \hline
    Waterfall bay & 170 & 4.65\%\\
    \hline
    Pufferfish & 110 & 3.01\%\\
    \hline
    Wetlands & 59 & 1.61\%\\
    \hline
    Amazon rapids & 78 & 2.13\%\\
    \hline
    Europe & 182 & 4.98\%\\
    \hline
    External & 212 & 5.80\%\\
    \hline
    Creeks & 116 & 3.17\%\\
    \hline
    Caiman & 85 & 2.33\%\\
    \hline
    Australian lake & 143 & 3.91\%\\
    \hline
    Mimicry & 116 & 3.17\%\\
    \hline
    Neotropic & 229 & 6.27\%\\
    \hline
    Oceania & 104 & 2.85\%\\
    \hline
    Cavefish & 81 & 2.22\%\\
    \hline
    Electric fish & 85 & 2.33\%\\
    \hline
    Flooded plain & 264 & 7.22\%\\
    \hline
    Flooded plain (flooding) & 233 & 6.38\%\\
    \hline
    Flooded plain (dry) & 95 & 2.60\%\\
    \hline
    Resurgences & 71 & 1.94\%\\
    \hline
    Rivers of Bonito & 413 & 11.30\%\\
    \hline
    Rivers of Pantanal & 172 & 4.71\%\\
    \hline
    Paths & 114 & 3.12\%\\
    \hline
\end{tabular}%
}
\end{table}

\subsection*{Neural Networks}

The task at hand is classifying tank images according to the thematic tanks. In the field of computer vision, the use of convolutional neural networks has achieved remarkable results. In the last few years, another similar but ultimately different kind of technology has become the object of interest in the field: vision transformers.

In 2017, the work of Vaswani \textit{et al.}~\cite{vaswani_attention_2017} in the field of natural language processing resulted in this new kind of neural network, the transformers, a smaller neural network that achieved state-of-the-art results by making use of an attention mechanism called \textit{scaled dot-product attention} and much fewer convolutional layers. In 2020, Dosovitskiy \textit{et al.}~\cite{dosovitskiy_an_image_2020} proposed a sort of ``pure'' transformer for computer vision tasks, precisely the Vision Transformer (ViT) . Recognizedly, their work was not the first in which the use of transformers was attempted in Computer Vision. However, as the authors claim, it was the first that showed that a transformer could be used in a ``pure'' state (\textit{i.e.}, without mixing a transformer-based architecture with a convolutional approach) and achieve state-of-the-art results.

Vision transformers have become a hot topic in Computer Vision ever since. New transformer-based architectures and new transformer-with-convolution-based architectures have been proposed in the last few years. Given that this is how things stand, the proposed experiment shall compare the performance of the following neural networks: Resnet18, MaxViT, LamHaloBotNet, LambdaResnet. Furthermore, given the proposal described in Section~\ref{proposal}, we also test and compare three relatively smaller networks: EfficientNet, DenseNet and MobileNet. All networks will be compared to their performance differences given their sizes, which will form the basis for deciding whether the networks should be processed online or on the users' smartphones.

Below, we comment briefly on each one of the architectures evaluated. Also, Table~\ref{table:num_params} shows the number of parameters of each network.

\begin{itemize}
    \item Resnet18\footnote{In this work, we use the PyTorch/Torchvision implementation.}: Residual networks were proposed by He \textit{et al.}~\cite{he_deep_2016} in 2015 as a way to deal with the fact that the increase in depth of a neural network eventually leads to a degradation in accuracy, when the network gets too deep. The main idea is to adjust the weights in a group of layers with reference to the input given to that group. This reference is accomplished by adding the input itself (as it originally is, or linearly transformed if so required by the dimensionality) to the output of the layers, which makes optimization easier.
    
    \item MaxViT\footnote{In this work, we tested the MaxViT-T as implemented in PyTorch Image Models~\cite{rw2019timm} (timm; baptized in implementation as maxvit\_rmlp\_tiny\_rw\_256), available here: https://github.com/rwightman/pytorch-image-models.}: Multi-Axis Vision Transformers were proposed in 2022 by Tu \textit{et al.}~\cite{tu_maxvit_2022}. In fact, the authors proposed, first and foremost, a transformer module called multi-axis self-attention, and what is here called MaxViT is the neural network, also proposed by the authors, composed by blocks constituted by the module and convolutional layers. It is, therefore, a hybrid architecture, both transformer- and convolution-based. In short, the fundamental idea of the new block is to use block attention, attention over windows of the input map, to get local information, and grid attention, attention diffused across smaller pieces distributed across the input map, to get global spatial information. The purpose is to improve the model capacity of the transformer, while keeping computational costs low.
    
    \item LambdaResnet and LamHaloBotNet\footnote{Both available here: https://github.com/rwightman/pytorch-image-models as well. Wightman~\cite{rw2019timm}, maintainer of the implementations, considers them ``experimental variants''.}: these are Neural Networks implemented in PyTorch Image Models in a ``bring-your-own-blocks'' fashion. They make use of various attention mechanisms. Lambda layers, as well as LambdaResnets, were proposed in 2021 by Bello \cite{bello_lambdanetworks_2021}. LambdaResnets are Resnets in which the bottleneck blocks were substituted by lambda layers. Lambda layers were proposed as a way to model contextual information out of the self-attention framework, as a way to reduce memory cost. HaloNets were proposed in 2021 by Vaswani \textit{et al.}~\cite{vaswani_scaling_2021}. The authors explore the similarities between self-attention and convolution, and propose to apply self-attention by getting queries from blocks established on the input, and keys and values from a window created around the query block, including this block itself. Bottleneck transformers are a kind of architectural design whose importance was pointed out by Srinivas \textit{et al.}~\cite{srinivas_bottleneck_2021}. In short, it consists in the possibility of replacing convolutions with the transformer's multi-head self-attention in certain cases, such as that of the bottleneck block of ResNets.

    \item EfficientNet: the family of EfficientNet networks was proposed by Tan and Le~\cite{tan2019efficientnet} as the result of a research on how to efficiently scale deep neural networks up. In their work, the authors proposed a compound scaling method to scale network depth, width and resolution in a uniform way. To test their method, they proposed a baseline EfficientNet-B0, along with efficiently scaled up versions from B1 up to B7. In this work, we evaluate the B2 version.

    \item MobileNetV3: the MobileNetV3 networks were proposed by Howard \textit{et al.}~\cite{howard2019searching}, and constitute the third generation of MobileNets. This family of networks was conceived within the ideal of optimizing neural networks for mobile devices. In this third generation, important changes were introduced, such as the redesign of layers and the use of the hard swish as non-linearity.

    \item DenseNet121\footnote{In this work, we used the Pytorch/Torchvision implementation.}: Densely Connected Convolutional Networks were introduced by Huang \textit{et al.}~\cite{huang2016densely}. In their work, the authors proposed a network design where the output feature maps of each layer are used as inputs to all subsequent layers, and all subsequent layers have as inputs the feature maps of all preceding layers, up until the input image. The objective was to improve the flow of information and gradient throughout the layers, by sharing the feature maps across the entire network. With this idea, the authors managed to achieve results that were close to the state-of-the-art with a smaller number of parameters.
\end{itemize}

\begin{table}[htb!]
\centering
\caption{Number of parameters of each neural network evaluated in this work.}
\label{table:num_params}
\begin{tabular}[t]{|c|c|}
    \hline
    Architecture & \# Parameters \\
    \hline
    Resnet18 & 11,689,512 \\
    \hline
    MaxViT & 29,148,896 \\
    \hline
    LambdaResnet & 10,988,688 \\
    \hline
    LamHaloBotNet & 22,569,824 \\
    \hline
    EfficientNet & 9,109,994 \\
    \hline
    MobileNetV3 & 5,483,032 \\
    \hline
    DenseNet121 & 7,978,856 \\
    \hline
\end{tabular}%
\end{table}

\subsection*{Experimental Setup}

The selected network architectures' weights were adjusted by the Adaptive Moment Estimation (Adam) optimizer. The combinations were tested using a stratified 10-fold cross-validation sampling strategy. In each fold, 20\% of the training data was used for validation. The maximum possible number of epochs was 1000. However, validation loss values were monitored for early stopping, with a tolerance of 0.1 and 30 epochs of patience. The architectures were initialized with the available pre-trained weights (transfer learning) in all cases. The images were resized to (256, 256) since this is the required image size for all the neural network implementations utilized. The following data augmentation techniques were used: random horizontal and vertical flips, with a probability of 0.5; random 90-degree rotation; random crop; and random perspective alterations.\footnote{Implemented in Torchvision as RandomHorizontalFlip, RandomVerticalFlip, RandomRotation, RandomCrop, and RandomPerspective.} The weights were optimized using a learning rate of 0.001. The training was performed in batches of 12 images.

The results were used to calculate statistics for three performance metrics: precision, recall, and f-score. A one-way ANOVA hypothesis test was used to compare the effects of architecture variation on each performance metric. The chosen significance level was 5\%. The Scott-Knott clustering test followed each ANOVA to specify the statistically different architectures. The ANOVA and the post-hoc Scott-Knott tests were conducted in the R programming language. For the ANOVA, the R Stats Package was used. The Scott-Knott clustering test was performed with the ScottKnott package \cite{jelihovschi2014scottknott}. Boxplots, confusion matrices and Receiver Operating Characteristic (ROC) curves were also produced and used in the analysis.

\section*{Results and Discussion}

\begin{figure*}[t]
    \centering
    \includegraphics[width=1\textwidth]{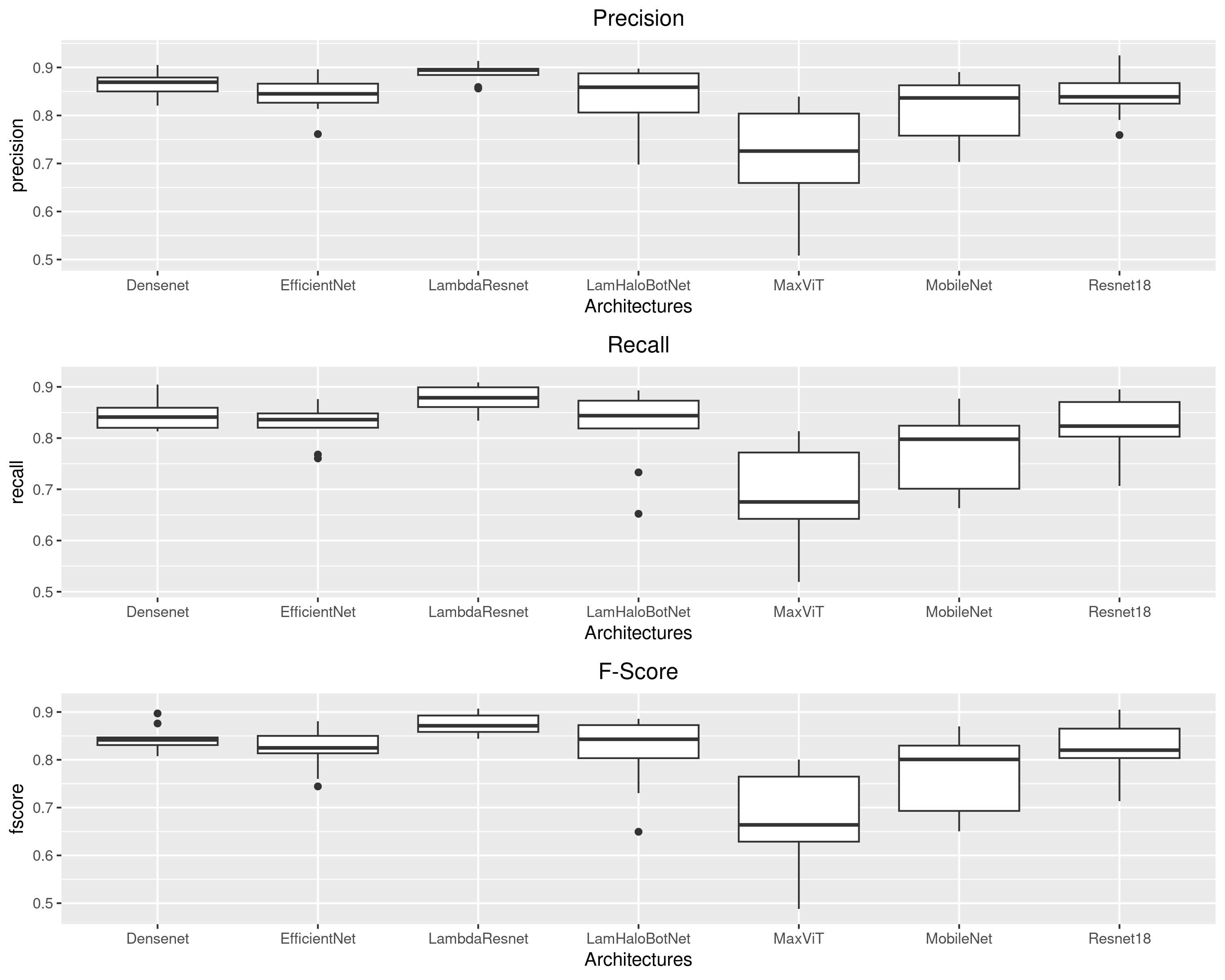}
    \caption{Boxplot for each architecture used in the experiment.}
    \label{fig:boxplot_10folds}
\end{figure*}

\begin{figure}[!htb]
    \centering
    \includegraphics[width=0.7\textwidth]{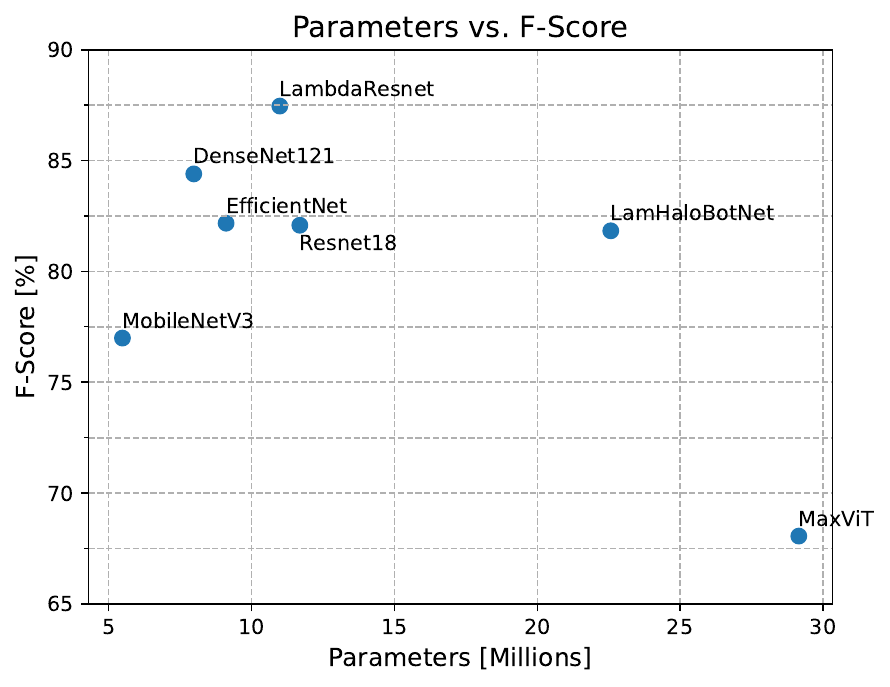}
    \caption{Number of parameters vs. f-score results for each architecture.}
    \label{fig:fscore_params}
\end{figure}

Figure \ref{fig:boxplot_10folds} shows boxplots for precision, recall, and f-score for each architecture across ten folds. Table \ref{table:precision} shows the median, interquartile range (IQR), mean, and standard deviation (SD) for precision, recall, and f-score for each architecture. Figure \ref{fig:fscore_params} compares the number of parameters of each neural network with the f-score results achieved by them. Considering the convex hull of the graph, MobileNetV3, DenseNet121, and LambdaResnet are the best trade-off of F-score and number of parameters.

{\renewcommand{\arraystretch}{1.2}
\begin{table*}[!htbp]
\centering
\caption{Precision, recall and f-score statistics for each architecture. Scott-Knott clustering test results are presented as capital letters next to the mean values.}
\label{table:precision}
\begin{tabular}[t]{|c|c|c|c|c|}
\hline
\multicolumn{5}{c}{\textbf{Precision}} \\
\hline
Architecture & Median & IQR & Mean & SD\\
\hline \hline
Resnet18 & 0.8388610 & 0.0427257 & 0.8431657 a & 0.0488431\\
\hline
MaxViT & 0.7258782 & 0.1445138 & 0.7181872 b & 0.1013066\\
\hline
LambdaResnet & 0.8943776 & 0.0129604 & 0.8891400 a & 0.0190454\\
\hline
LamHaloBotNet & 0.8587547 & 0.0815401 & 0.8371838 a & 0.0646207\\
\hline
EfficientNet & 0.8450530 & 0.0396771 & 0.8416580 a & 0.0376566\\
\hline
MobileNetV3 & 0.8364602 & 0.1048045 & 0.8118989 a & 0.0654814\\
\hline
DenseNet121 & 0.8692268 & 0.0290350 & 0.8665466 a & 0.0268982\\
\hline

\multicolumn{5}{c}{\textbf{Recall}} \\
\hline
Architecture & Median & IQR & Mean & SD\\
\hline \hline
Resnet18 & 0.8234125 & 0.0676031 & 0.8222047 a & 0.0594642\\
\hline
MaxViT & 0.6753436 & 0.1295495 & 0.6924280 c & 0.0916703\\
\hline
LambdaResnet & 0.8787862 & 0.0385103 & 0.8775279 a & 0.0253155\\
\hline
LamHaloBotNet & 0.8438065 & 0.0541796 & 0.8224366 a & 0.0750914\\
\hline
EfficientNet & 0.8360617 & 0.0279304 & 0.8268298 a & 0.0367740\\
\hline
MobileNetV3 & 0.7977565 & 0.1230585 & 0.7743310 b & 0.0786110\\
\hline
DenseNet121 & 0.8410389 & 0.0391698 & 0.8431819 a & 0.0286452\\
\hline

\multicolumn{5}{c}{\textbf{F-score}} \\
\hline
Architecture & Median & IQR & Mean & SD\\
\hline \hline
Resnet18 & 0.8201954 & 0.0615506 & 0.8208287 a & 0.0614754\\
\hline
MaxViT & 0.6638640 & 0.1360363 & 0.6806592 b & 0.0962083\\
\hline
LambdaResnet & 0.8712313 & 0.0343017 & 0.8745617 a & 0.0212171\\
\hline
LamHaloBotNet & 0.8429809 & 0.0691919 & 0.8183138 a & 0.0762315\\
\hline
EfficientNet & 0.8249585 & 0.0364336 & 0.8217447 a & 0.0419911\\
\hline
MobileNetV3 & 0.8007643 & 0.1366063 & 0.7699452 a & 0.0809824\\
\hline
DenseNet121 & 0.8415972 & 0.0158589 & 0.8439717 a & 0.0258583\\
\hline
\hline
\end{tabular}

\end{table*}
}

The ANOVA test resulted in highly significant results, with \textit{p}-values of $4.9 \cdot 10^{-7}$ for precision, $2.46 \cdot 10^{-7}$ for recall and $1.31 \cdot 10^{-7}$ for f-score. Also, in Table~\ref{table:precision}, the results of the Scott-Knott clustering test applied for each metric are presented next to the mean values. The architectures with the same letters were clustered and are not statistically different. 

\begin{figure}[!htb]
    \centering
    \includegraphics[width=0.7\columnwidth]{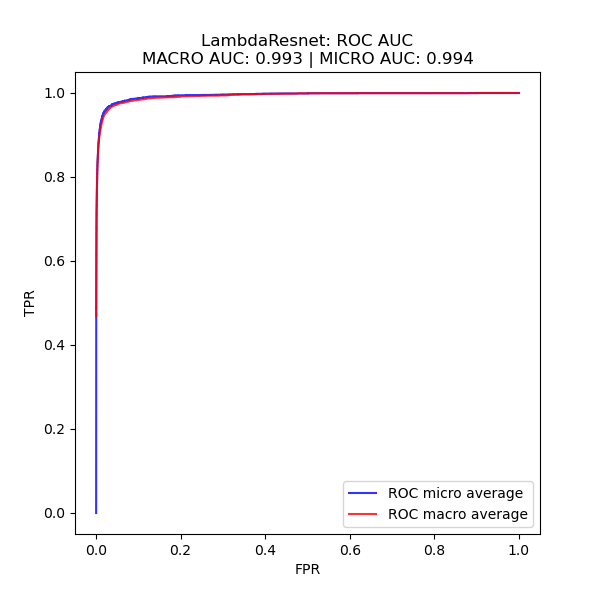}
    \caption{ROC curves calculated as micro and macro averages for the LambdaResnet architecture.}
    \label{fig:roc_lambdaresnet}
\end{figure}

\begin{figure}[!htb]
    \centering
    \includegraphics[width=0.7\columnwidth]{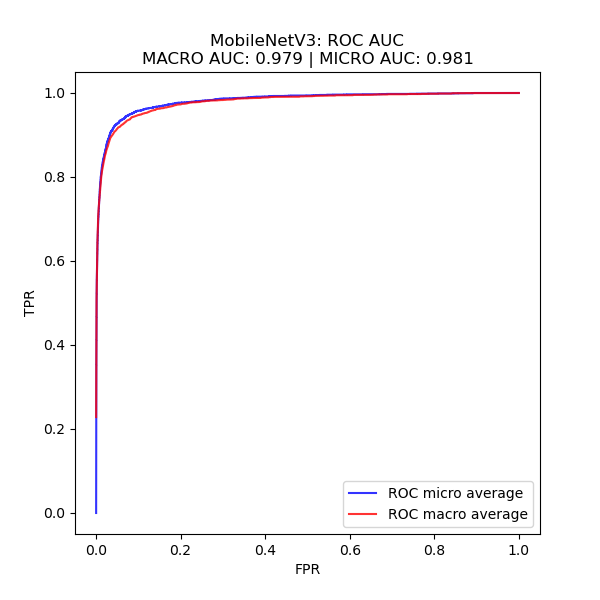}
    \caption{ROC curves calculated as micro and macro averages for the MobileNetV3 architecture.}
    \label{fig:roc_mobilenet}
\end{figure}

Figures~\ref{fig:roc_lambdaresnet} and~\ref{fig:roc_mobilenet} show the Receiver Operating Characteristic (ROC) curves for LambdaResnet (11 million parameters), which achieved the highest average performance, as it is possible to see in Table~\ref{table:precision}, and for the MobileNetV3 architecture (5.4 million parameters), which is the smallest architecture evaluated in this work, as can be seen in Table~\ref{table:num_params}. These curves were calculated with a One-vs-Rest strategy, which means that the number of classes and the fact that they are imbalanced may have affected the results. Nonetheless, the ROC curves present an AUC higher than 0.98, indicating that the results of f-score can be even higher when using different classification thresholds. In both cases, we can see a steady increase in FPR, which allows us to set a classification threshold to remove false positives from the localization.

Two last observations regarding the dataset. First, in section~\ref{methods}, we said that the students who captured the images were instructed to keep the dataset balanced. However, this did not happen, as seen in table~\ref{table:tanks}. According to the students themselves, this happened because some fish were too small, and some students insisted on taking a picture of such fish. In order to do so, they took more pictures of tanks with such small fish than of others. Therefore, the imbalance in the dataset can be understood as a characteristic of the problem reflected in the dataset. Finally, the natural behavior of the students also caused the dataset to have some sets of images that are too similar to each other, which may affect the results, but is also an aspect of the problem.

\subsection*{Conclusion}
This work was motivated by the possibility of improving the experience of learning about aquatic environments by visiting the Pantanal Biopark, the largest freshwater aquarium in the world. As reported, the task of using AI to improve visiting experience in cultural contexts is not without precedent, and in this work it was our intention to give an important step towards such a goal in the context of the Biopark. More specifically, in this work we presented a new dataset composed by images of many of the tanks present in the Biopark. The task that comes with the dataset is to correctly identify the tanks from which the images were taken. By applying some state-of-the-art neural network architecture, we established some baseline results of around 88\% for precision, recall and f-score. Although not bad, these results are still not ideal to build a system with the proposed objectives. As discussed, it is our opinion that there is room for improvement, both by testing other architectures and optimizers, and also by testing different hyperparameters for the same architectures and for the Adam optimizers. This question is left open for future research.

\section*{Acknowledgments}
This work has received financial support from the Dom Bosco Catholic University and from the Foundation for the Support and Development of Education, Science and Technology from the State of Mato Grosso do Sul, FUNDECT. Some of the authors have been awarded with Scholarships from the the Brazilian National Council of Technological and Scientific Development, CNPq and the Coordination for the Improvement of Higher Education Personnel, CAPES. We thank Biopark Pantanal for supporting this work.

We also thank the following students for collecting the images: Álvaro Barbosa, Amanda Yamashita, Dalian Gowert, Fernanda Cacho, Guilherme Cândido, Guilherme Lube, João Pedro Roberto, Leonardo Victorio, and Rodrigo Stuqui. \cite{bib1}


%
%
%


%
%
%
%

\end{document}